\documentclass{article}
\usepackage{ijcai25}

\usepackage{times}
\usepackage{soul}
\usepackage{url}
\usepackage[hidelinks]{hyperref}
\usepackage[utf8]{inputenc}
\usepackage[small]{caption}
\usepackage{graphicx}
\usepackage{amsmath}
\usepackage{amsthm}
\usepackage{booktabs}
\usepackage{algorithm}
\usepackage{algorithmic}
\usepackage[switch]{lineno}

\usepackage{threeparttable}
\usepackage{subfigure}
\usepackage{multirow}
\usepackage{amssymb}
\usepackage{enumitem}
\usepackage{ragged2e}

\usepackage{hyperref}
\usepackage{orcidlink}
\usepackage[misc]{ifsym}

\title{Multi-Sourced Compositional Generalization in Visual Question Answering}

\author{
Chuanhao Li$^{2,1}$\textsuperscript{*}\and
Wenbo Ye$^{2,1}$\textsuperscript{*}\and
Zhen Li$^{1}$\and
Yuwei Wu$^{1,2}$\textsuperscript{$\dagger$}\and
Yunde Jia$^{2,1}$\\
\affiliations
$^1$Beijing Key Laboratory of Intelligent Information Technology,\\School of Computer Science \& Technology, Beijing Institute of Technology, China\\
$^2$Guangdong Laboratory of Machine Perception and Intelligent Computing,\\Shenzhen MSU-BIT University, China\\
\emails
\{lichuanhao, yewenbo, li.zhen, wuyuwei, jiayunde\}@bit.edu.cn
}

\begin{document}

\maketitle
\renewcommand{\thefootnote}{\fnsymbol{footnote}}
{\let\thefootnote\relax\footnotetext{
\noindent \hspace{-5mm}
\textsuperscript{*} Equal contribution;
\textsuperscript{$\dagger$} Corresponding author: Yuwei Wu.}}

\begin{abstract}
Compositional generalization is the ability of generalizing novel compositions from seen primitives, and has received much attention in vision-and-language (V\&L) recently.
Due to the multi-modal nature of V\&L tasks, the primitives composing compositions source from different modalities, resulting in multi-sourced novel compositions.
However, the generalization ability over multi-sourced novel compositions, \textit{i.e.}, multi-sourced compositional generalization (MSCG) remains unexplored.
In this paper, we explore MSCG in the context of visual question answering (VQA),
and propose a retrieval-augmented training framework to enhance the MSCG ability of VQA models by learning unified representations for primitives from different modalities.
Specifically, semantically equivalent primitives are retrieved for each primitive in the training samples, and the retrieved features are aggregated with the original primitive to refine the model.
This process helps the model learn consistent representations for the same semantic primitives across different modalities.
To evaluate the MSCG ability of VQA models, we construct a new GQA-MSCG dataset based on the GQA dataset, in which samples include three types of novel compositions composed of primitives from different modalities.
Experimental results demonstrate the effectiveness of the proposed framework.
We release GQA-MSCG at \href{}{https://github.com/NeverMoreLCH/MSCG}.
\end{abstract}

\section{Introduction}

\begin{figure}[t]
    \centering
    \includegraphics[width=1\linewidth]{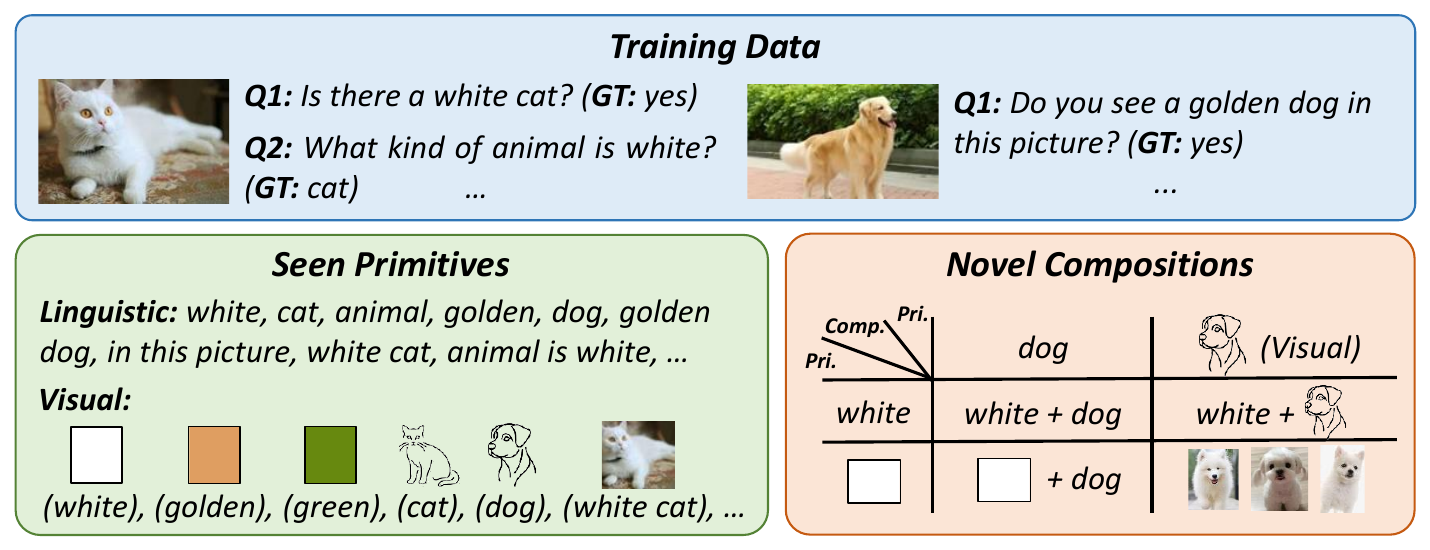}
    \caption{Multi-sourced novel compositions in the context of VQA.}
    \label{fig:motivation_mscg}
\end{figure}

Compositional generalization refers to the ability of generalizing novel compositions from seen primitives.
In vision-and-language (V\&L), the primitives of a composition come from either the linguistic modality or the visual modality, \textit{i.e.}, compositions are multi-sourced.
As shown in Figure \ref{fig:motivation_mscg}, in the context of visual question answering (VQA), “white (linguistic modality) + dog (linguistic modality)” is an novel composition, and “white (linguistic modality) + \raisebox{-0.5ex}{\includegraphics[height=4mm]{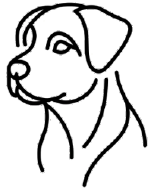}} (visual modality)” and “\raisebox{-0.5ex}{\includegraphics[height=4mm]{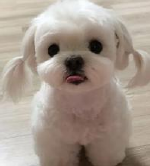}} (white + dog in visual modality)” are also novel compositions.
However, prior works \cite{dankers2022paradox,li2024context,li2023variational} only consider novel compositions of primitives from a single modality (\textit{e.g.}, “white + dog”).
Whether the model's generalization ability to different modality primitives' novel compositions (\textit{i.e.}, multi-sourced novel compositions) remains unexplored.
To generalize over multi-sourced compositions, a model needs to not only have the ability to understand individual primitives but also the ability to align primitives of different modalities.

To address the above issue, we propose a retrieval-augmented training framework.
The basic idea of the framework is to learn similar representations for primitives from different modalities by retrieving semantically equivalent primitives, thereby maintaining consistent generalization ability for multi-sourced compositions.
Specifically, the framework consists of three key components: retrieval database construction, feature retrieval, and feature aggregation.
For the retrieval database, we construct separate primitive databases for the linguistic and visual modalities, where words with the same prototype are treated as the same linguistic primitive, and visual entities with the same label are treated as the same visual primitive. 
For the same primitive, both the linguistic/visual primitive databases contain multiple instances of that primitive in different contexts.
For example, for the linguistic primitive “\textbf{dog}”, the linguistic primitive database contains instances of 
“Is the \textbf{dog} black?”, 
“How many \textbf{dogs} are there?”, 
“Do you see a golden \textbf{dog} in this picture?”
in different contexts.
During training, for each primitive in the training sample, semantically similar primitives from both the linguistic and visual primitive databases are retrieved.
The retrieved features are then aggregated with the original primitive features to optimize the model.
Since the representations in the primitive databases are updated throughout the training, the model continuously refines its current representation by leveraging the retrieved relevant features to learn similar representations for the same semantic primitives across different modalities and contexts.

To quantitatively evaluate the generalization ability of VQA models for multi-sourced novel compositions, we construct the GQA-MSCG dataset based on the GQA dataset \cite{cvpr2019gqa}.
We categorize compositions based on the modality of the primitives from the composition, resulting in three types of novel compositions: [Linguistic primitive, Linguistic primitive] (LL), [Visual primitive, Visual primitive] (VV), and [Linguistic primitive, Visual primitive] (LV).
We construct three basic test splits, labeled $LL$, $VV$, and $LV$, each containing test samples of different types of novel compositions.
To further explore how the co-occurrence of different types of novel compositions in samples affects model performance, we construct test splits, where each sample simultaneously contain two types of novel compositions: $LL+VV$, $LL+LV$, and $VV+LV$, as well as a test split, where each sample contain all three types of novel compositions: $LL+VV+LV$.
Experimental results demonstrate that the proposed framework significantly improves VQA models' generalization ability to multi-sourced novel compositions while maintaining their independent and identically distributed (IID) generalization ability.

To sum up, our contributions are as follows:
\begin{itemize}
    \item We are the first to explore the multi-sourced compositional generalization in V\&L, which is critical for cross-modal understanding.
    \item We propose a retrieval-augmented training framework that improves the multi-sourced compositional generalization ability of VQA models by learning similar representations for primitives from different modalities.
    \item We present a GQA-MSCG dataset to evaluate the multi-sourced compositional generalization ability of VQA models with different types of novel compositions.
\end{itemize}

\section{Related Work}

Compositional generalization has garnered significant attention in various research fields.
In natural language processing (NLP), works \cite{li2021compositional,dankers2022paradox} focus on improving the compositional generalization ability of models for tasks like machine translation.
Additionally, Chai et al. \shortcite{chai2024compositional} used text generation models for data augmentation during training to enhance compositional generalization in multi-label text classification tasks.
In computer vision (CV), works \cite{naeem2021learning,jing2024retrieval,li2024context} focus on improving compositional generalization for tasks like compositional zero-shot learning, specifically for novel compositions of [visual primitive, visual primitive].
In V\&L, Pantazopoulos et al. \shortcite{cg1} constructed the first dataset to evaluate the compositional generalization ability of image captioning models.
Works \cite{bahdanau2018systematic,neurips2021lgnmn,yamada2024transformer,li2023exploring} focus on enhancing the compositional generalization ability of VQA models.
Moreover, Li et al. \shortcite{cvpr2022visa} constructed the Charades-CG and ActivityNet-CG datasets for temporal video grounding (TVG) and used variational cross-graph reasoning to improve the compositional generalization ability of TVG models.
These works primarily focus on improving the generalization ability of models for novel compositions involving the same modality of primitives, \textit{e.g.}, works in NLP and V\&L focus on [linguistic primitive, linguistic primitive] compositions, and works in CV focus on [visual primitive, visual primitive] compositions.
In contrast, we explore the compositional generalization ability of V\&L models for novel compositions involving different modality primitives, construct a new benchmark GQA-MSCG, and propose a retrieval-augmented training framework to improve the model's generalization ability for different types of novel compositions.

\section{Framework}

\begin{figure*}[tb]
	\centering
	\includegraphics[width=0.97\linewidth]{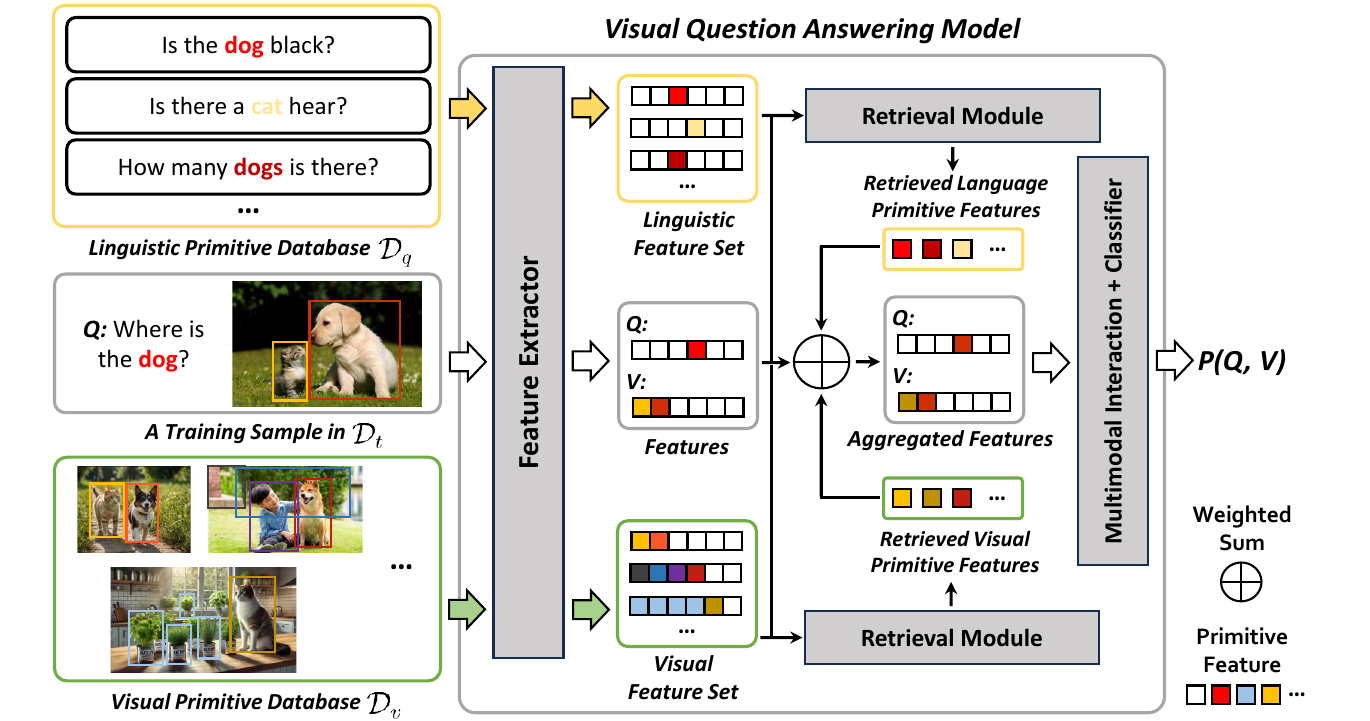}
	\caption{The overall framework of the proposed framework.}
	\label{fig:framework_mscg}
\end{figure*}

\subsection{Overview}

The overall framework of the proposed framework in the context of VQA is shown in Figure \ref{fig:framework_mscg}.
For the training set $\mathcal{D}_t$, the first step is to construct linguistic and visual primitive databases, $\mathcal{D}_q$ and $\mathcal{D}_v$, respectively, for retrieval purposes.
primitives with the same category (e.g., dog, cat) or attribute (e.g., blue, big) are treated as the same type of primitive.
Both $\mathcal{D}_q$ and $\mathcal{D}_v$ contain multiple instances of each primitive in different contexts from $\mathcal{D}_t$.
During training, for each training sample $(Q, V)$, where $Q$ represents the question and $V$ represents the image, the retrieval module is used to retrieve similar primitives from $\mathcal{D}_q$ and $\mathcal{D}_v$ for the primitives in $Q$ and $V$ at feature level, respectively.
The retrieved features are then aggregated with the original primitive features and used to replace the original features for training.
Through this training process, the VQA model continuously refines its feature extractor by leveraging the retrieved relevant features, thereby learning similar features for the same semantic primitives across different modalities and contexts, thus improving the generalization ability of the model for multi-sourced novel compositions.

\subsection{Retrieval Database Construction}
\label{sec:db_construct}

To ensure fairness in the comparison, the retrieval database is constructed based on the training set $\mathcal{D}_t$ to avoid introducing external data during training.

\noindent
\textbf{Linguistic Primitive Database.} 
For the linguistic primitive database, we first extract all words from the questions in $\mathcal{D}_t$ using the NLTK toolkit \cite{bird2009natural}.
All words are lemmatized, and words with part-of-speech tags as nouns, verbs, adjectives, and adverbs are considered as linguistic primitives.
The set of all unique linguistic primitives is denoted as $\mathcal{S}_q$.
For each linguistic primitive, $T_q$ questions containing the primitive are sampled from $\mathcal{D}_t$.
Although the meaning of the primitive remains the same across these $T_q$ questions, the different questions provide different contexts.
In different contexts, the features of the same linguistic primitive, extracted using a recurrent neural network, will differ.
Thus, it is necessary to sample different questions for each linguistic primitive.
To preserve the different contexts of the linguistic primitives, the linguistic primitive database stores the original questions, not just the linguistic primitive itself.
The resulting linguistic primitive database is represented as $\mathcal{D}_q = \bigcup_{p \in \mathcal{S}_q} f_q(p)$,
where $f_q(p) = \{ p_i \}_{i=1}^{T_q}$ represents the $T_q$ sampled questions containing the linguistic primitive $p$.

\noindent
\textbf{Visual Primitive Database.}
The process of constructing the visual primitive database relies on the scene graph annotations, which contains information about object categories and attributes in the images. 
Objects in the images are treated as visual primitives, with category and attribute information serving as the labels of these primitives.
After removing duplicates from the images in $\mathcal{D}_t$, the resulting set of labels is denoted as $\mathcal{S}_v$.
Similar to linguistic primitives, even though different visual primitives may share the same label, they represent different visual expressions of objects with the same type/attribute.
Therefore, for each label, $T_v$ images containing the visual primitive with the label are sampled.
Since different VQA models use different visual encoders, the input forms are not limited to object-level features.
To address this, the visual primitive database stores the original images, not just the visual primitives.
The resulting visual primitive database is represented as $\mathcal{D}_v = \bigcup_{l \in \mathcal{S}_v} f_v(l)$,
where $f_v(l) = \{ p_i^{(l)} \}_{i=1}^{T_v}$ represents the $T_v$ sampled images containing visual primitives with label $l$.

\subsection{Feature Retrieval and Aggregation}

To perform retrieval and aggregation at the feature level, both the training samples and all primitives in the two primitive databases ($\mathcal{D}_q$ and $\mathcal{D}_v$) need to pass through the feature extractors of the VQA model.
For a question $Q$, we obtain its feature $\textbf{\textit{h}}_q = g_q(Q) = \{ \textbf{\textit{h}}_q^i \}_{i=1}^{N}$, where $g_q(\cdot)$ denotes the linguistic feature extractor of the VQA model.
Here, $\textbf{\textit{h}}_q^i$ represents the feature of the $i$-th word, and $N$ is the number of words.
For an image $V$, the visual feature extractor $g_v(\cdot)$ typically produces two types of features: ``object-level" features or ``patch-level" features,
which can be represented as $\textbf{\textit{h}}_v = g_v(V) = \{ \textbf{\textit{h}}_v^i \}_{i=1}^{M}$,
where $\textbf{\textit{h}}_v^i$ represents the feature of the $i$-th object/patch, and $M$ is the number of objects/patches.

After obtaining the primitive-level features (word-level, object-level, patch-level),
for each primitive feature in the training sample, retrieval is performed on the primitive feature sets of the questions and images in $\mathcal{D}_q$ and $\mathcal{D}_v$ (processed by $g_q(\cdot)$ and $g_v(\cdot)$).
Specifically, for a primitive feature $\textbf{\textit{p}}$ in the training sample,
the top $K_q$ most similar primitive features $\{ \textbf{\textit{p}}_q^{(i)} \}_{i=1}^{K_q}$ are retrieved from the primitive feature set in $\mathcal{D}_q$,
and the top $K_v$ most similar primitive features $\{ \textbf{\textit{p}}_v^{(i)} \}_{i=1}^{K_v}$ are retrieved from the primitive feature set in $\mathcal{D}_v$.
We use cosine similarity $\textup{cos}(\cdot, \cdot)$ to measure the similarity between two primitive features.
The features are aggregated using a weighted average
\begin{equation}
\textbf{\textit{p}}_a = \textbf{\textit{p}} + 
w_q \cdot \frac{\sum_{i=1}^{K_q} \textup{cos}(\textbf{\textit{p}}, \textbf{\textit{p}}_q^{(i)})}{K_q} + 
w_v \cdot \frac{\sum_{i=1}^{K_v} \textup{cos}(\textbf{\textit{p}}, \textbf{\textit{p}}_v^{(i)})}{K_v}, 
\end{equation}
where the hyperparameters $w_q$ and $w_v$ control the contributions of the different modality primitive databases.
We use the aggregated feature $\textbf{\textit{p}}_a$ to replace the original primitive feature $\textbf{\textit{p}}$ for training.

\subsection{Optimization}

The proposed framework is applicable to different VQA baseline models, using the same optimization approach as the baseline model without introducing additional training losses or constraints.
Therefore, for a VQA baseline model using the training loss $\mathcal{L}$,
for a training sample $(Q, V)$ with ground-truth $A$,
the training loss for both the baseline model and the model with our framework is the same:
\begin{equation}
\mathcal{L} = \textup{loss}(P(Q, V), A),
\end{equation}
where \( P(Q, V) \) represents the output of the VQA model (\textit{e.g.}, a distribution vector over the number of categories), and \( \textup{loss}(\cdot, \cdot) \) represents the training loss function, such as the cross-entropy loss used in the UpDn model \cite{anderson2018bottom}.

\section{GQA-MSCG Dataset}

In order to quantitatively evaluate the generalization ability of VQA models to multi-sourced novel compositions, we construct the GQA-MSCG dataset based on the GQA dataset \cite{cvpr2019gqa}.
The process of constructing the GQA-MSCG dataset consists of three main steps: composition extraction, sample filtering, and sample classification.

\noindent
\textbf{Composition Extraction.}
We use the same steps as in Section \ref{sec:db_construct} to extract the linguistic and visual primitives for all samples in the train\_balanced split $\mathcal{D}_{t}$ of the GQA dataset.
The linguistic and visual primitives for a sample $s \in \mathcal{D}_{t}$ are represented by $\mathcal{P}_{s}^{q}=\{p_{s}^{q_i}\}_{i=1}^{N}$ and $\mathcal{P}_{s}^{v}=\{p_{s}^{v_i}\}_{i=1}^{M}$, respectively,
where $N$ and $M$ denote the number of linguistic and visual primitives in the sample.
The compositions in sample $s$ are represented as $\mathcal{C}_{s}=\{ [p_1, p_2] | p_1 \in \mathcal{P}_{s}^{q} \cup \mathcal{P}_{s}^{v}, p_2 \in \mathcal{P}_{s}^{q} \cup \mathcal{P}_{s}^{v}, p_1 \neq p_2 \}$.
Thus, the complete set of primitives in $\mathcal{D}_{t}$ can be expressed as
$\mathcal{P}_{\mathcal{D}_{t}} = \bigcup_{s \in \mathcal{D}_{t}} \mathcal{P}_{s}^{q} \cup \mathcal{P}_{s}^{v}$,
and the complete set of compositions can be expressed as
$\mathcal{C}_{\mathcal{D}_{t}} = \bigcup_{s \in \mathcal{D}_{t}} \mathcal{C}_{s}$.
Similarly, the compositions in the val\_all split $\mathcal{D}_{v}$ of the GQA dataset can be processed to obtain all compositions.
We denote the set of all compositions as $\mathcal{C}_{\mathcal{D}_{v}}$.

\noindent
\textbf{Sample Filtering.}
For a sample $s \in \mathcal{D}_{v}$, if the sample satisfies the following conditions simultaneously, it is added to the candidate test sample set $\mathcal{D}_{c}$:
\begin{itemize}
    \item All primitives in the sample must appear in the training set — $\forall p \in \mathcal{P}_{s}^{q} \cup \mathcal{P}_{s}^{v}, p \in \mathcal{P}_{\mathcal{D}_{t}}$.
    \item The sample must contain at least one composition not seen in the training set (i.e., an novel composition) — $\exists c \in \mathcal{C}_{s}, c \notin \mathcal{C}_{\mathcal{D}_{t}}$.
\end{itemize}

\noindent
\textbf{Sample Classification.}
Based on the modality of the primitives in a composition, we categorize three types of novel compositions: [Linguistic primitive, Linguistic primitive] (LL),
[Visual primitive, Visual primitive] (VV), and [Linguistic primitive, Visual primitive] (LV).
Test samples in $\mathcal{D}_{c}$ are classified into seven categories:
(1) $LL$: Samples containing only novel compositions of the type LL.
(2) $VV$: Samples containing only novel compositions of the type VV.
(3) $LV$: Samples containing only novel compositions of the type LV.
(4) $LL+VV$: Samples containing both novel compositions of the type LL and VV.
(5) $LL+LV$: Samples containing both novel compositions of the type LL and LV.
(6) $VV+LV$: Samples containing both novel compositions of the type VV and LV.
(7) $LL+VV+LV$: Samples containing novel compositions of all three types: LL, VV, and LV.
Samples in $LL$, $VV$ and $LV$ are used to evaluate the multi-sourced compositional generalization ability,
while samples in $LL+VV$, $LL+LV$, $VV+LV$ and $LL+VV+LV$ are used to further evaluate the impact of the co-occurrence of different types of novel compositions on model performance.

\begin{figure}[tb]
	\centering
	\includegraphics[width=1\linewidth]{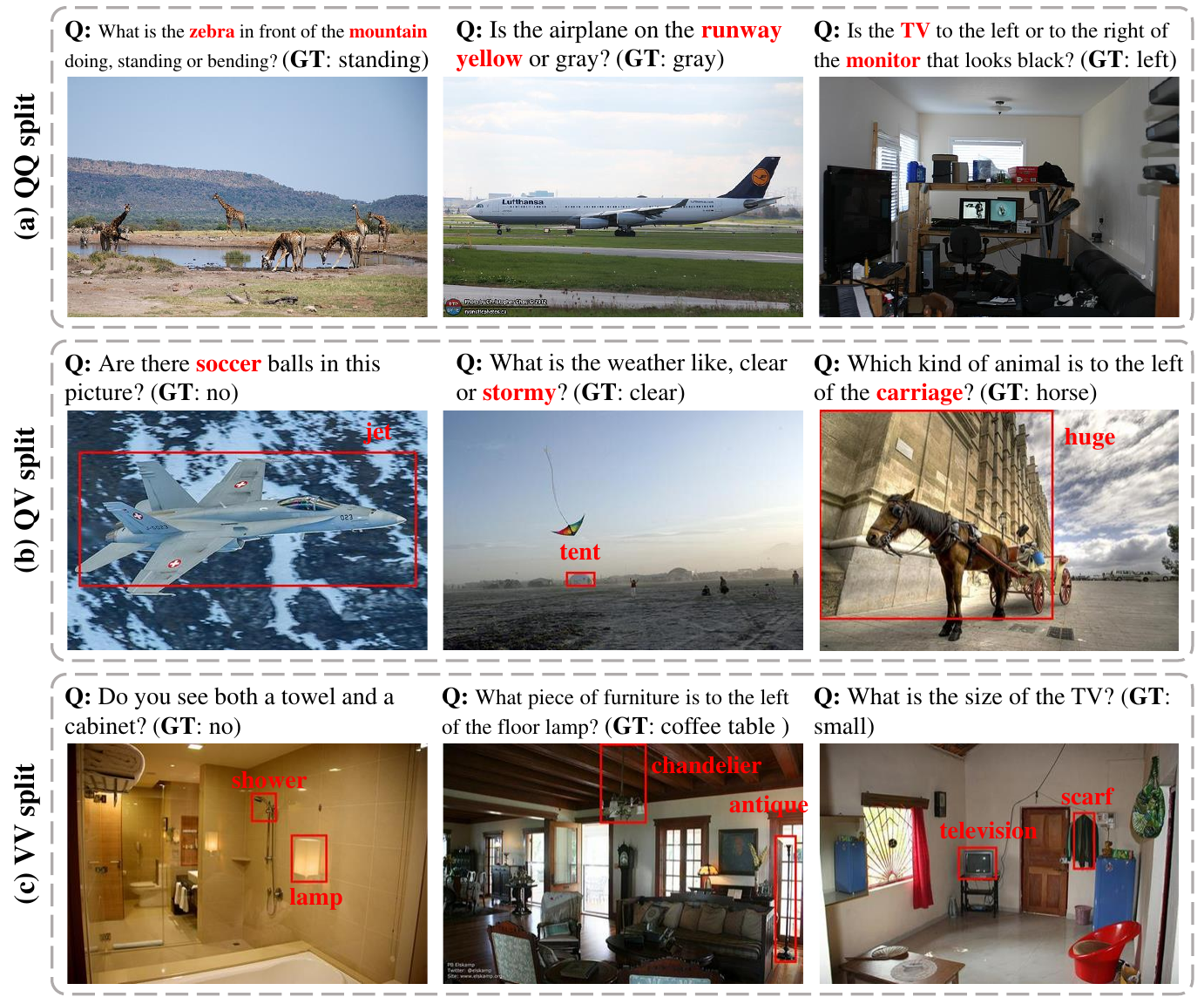}
	\caption{Level-1 samples in the GQA-MSCG dataset.}
	\label{fig:dataset1_mscg}
\end{figure}

\begin{figure}[!t]
	\centering
	\includegraphics[width=1\linewidth]{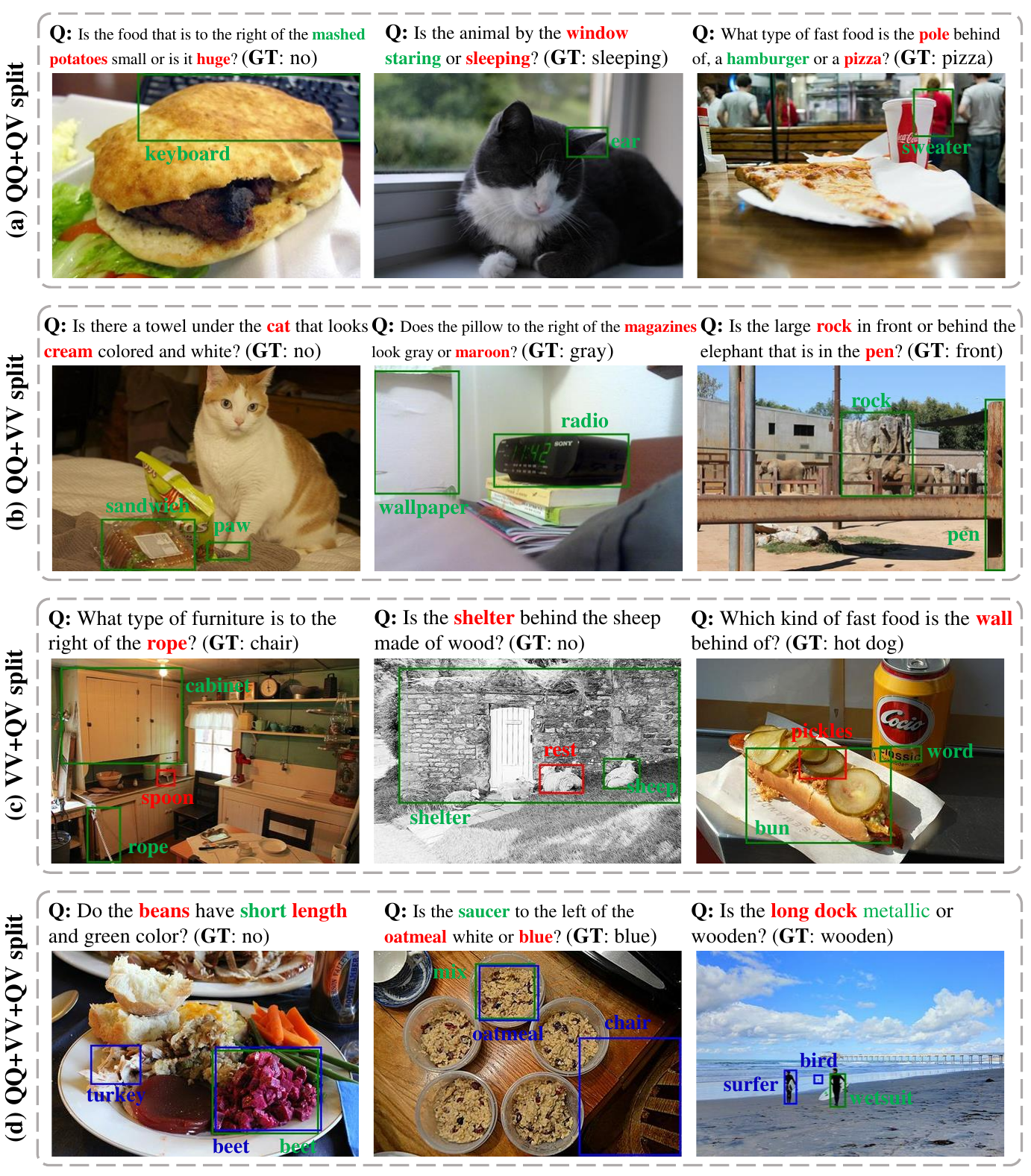}
	\caption{Level-2 samples and Level-3 samples in the GQA-MSCG dataset.}
	\label{fig:dataset2_mscg}
\end{figure}

\begin{table*}[!t]
\small
\caption{Accuracy (\%) on the GQA-MSCG dataset.}
\label{tab:gqamscg_mscg}
\centering
\begin{threeparttable}[b]
\setlength{\tabcolsep}{2.0mm}{
\begin{tabular}{clcccccccc}
\hline\noalign{\smallskip}
\multirow{2.5}{*}{Type} & \multirow{2.5}{*}{Model} & \multicolumn{3}{c}{Level-1} & \multicolumn{3}{c}{Level-2} & \multirow{1}{*}{Level-3} & \multirow{2.5}{*}{Overall} \\ 
\cmidrule(lr){3-5}\cmidrule(lr){6-8}\cmidrule(lr){9-9}
& & \textit{LL} & \textit{VV} & \textit{LV} & \textit{LL+VV} & \textit{LL+LV} & \textit{VV+LV} & \textit{LL+VV+LV} &\\
\cmidrule(){1-10}
\multirow{1}{*}{Small Model} & CFR \cite{nguyen2022coarse} & 74.78 & 72.18 & 73.36 & 72.26 & 73.56 & 70.94 & 69.78 & 72.41 \\
\multirow{1}{*}{($\leq$ 0.2B)} & \textbf{+ RAG (Ours)} & \textbf{76.28} & \textbf{73.88} & \textbf{75.64} & \textbf{73.58} & \textbf{75.66} & \textbf{73.18} & \textbf{71.22} & \textbf{74.21} \\
\cmidrule(){1-10}
\multirow{3.5}{*}{Large Model} & LLaVA-1.5 \cite{liu2023improvedllava}& 70.22 & 70.34 & 69.62 & 67.04 & 68.84 & 68.40 & 64.14 & 68.37 \\
\multirow{3.5}{*}{($\geq$ 7B)} & LLaVA-1.6 \cite{liu2024llavanext} & 72.36 & 71.52 & 72.60 & 69.70 & 69.50 & 69.98 & 67.60 & 70.46 \\
\cmidrule(){2-10}
& Qwen-VL \cite{Qwen-VL} & 72.24 & 65.54 & 69.06 & 68.18 & 72.08 & 66.04 & 66.32 & 68.49 \\
& \textbf{+ RAG (Ours)}& \textbf{74.68} & \textbf{68.90} & \textbf{71.98} & \textbf{71.50} & \textbf{74.82} & \textbf{68.92} & \textbf{69.88} & \textbf{71.53} \\
\noalign{\smallskip}\hline
\end{tabular}}
\end{threeparttable}
\end{table*}

For each category of test samples, we randomly sample 5,000 samples from $\mathcal{D}_{c}$, resulting in a total of 35,000 samples for the GQA-MSCG dataset.
Samples containing $x$ types of novel compositions are referred to as Level-$x$ samples, and the difficulty of the samples increases as $x$ increases.
For example, samples in $LL$, $VV$ and $LV$ are Level-$1$ samples, $LL+VV$, $LL+LV$ and $VV+LV$ are Level-$2$ samples, $LL+VV+LV$ are Level-$3$ samples.
As the level increases, the co-occurrence count of novel composition categories in the test samples continues to rise, making the difficulty of the test samples increasingly higher.
The test samples of GQA-MSCG are shown in Figure \ref{fig:dataset1_mscg} and Figure \ref{fig:dataset2_mscg}, where the words or visual regions of the same color in a sample represent a pair of novel compositions in the sample.

\section{Experiments}

\subsection{Experimental Setup}

\noindent
\textbf{Dataset.}
We evaluate proposed frameworks on the VQA task, three datasets are selected to validate the effectiveness of the proposed frameworks: the GQA dataset \cite{cvpr2019gqa}, the VQA v2 dataset \cite{goyal2017making} and our GQA-MSCG dataset. 
The GQA dataset is a widely used large-scale dataset in VQA, containing a large number of template-based compositional questions.
The VQA v2 dataset is an extended balanced version of the VQA v1 dataset \cite{vqa_first}, where questions are human-made.
These two datasets are often used to test the generalization ability of VQA models to IID data as well as their ability for compositional reasoning. 
Our GQA-MSCG dataset is used to evaluate the VQA model's ability to generalize consistently to multi-sourced novel compositions, which refers to the VQA model's ability to generalize to novel compositions of primitives from different modalities.

\noindent
\textbf{Baseline Models.}
We use CFR \cite{nguyen2022coarse} and Qwen-VL \cite{Qwen-VL} as baseline models.
The baseline models combined with the proposed framework are referred to as CFR+RAG and Qwen-VL+RAG, respectively.
CFR is a representative small model in the VQA task with a parameter size less than 0.2B.
Qwen-VL is a popular open-sourced multimodal large model (with more than 7B parameters), which can be applied to various vision-and-language tasks such as VQA, image captioning, visual dialogue, and others.
For both CFR and Qwen-VL, we reimplemented them based on their officially released code.

\noindent
\textbf{Implementation Details.}
For experiments on all three datasets including GQA, GQA-MSCG and VQA v2, 
we fine-tune Qwen-VL and Qwen-VL+RAG with LoRA \cite{hu2022lora} with a maximum of 2 epochs.
For CFR+RAG and Qwen-VL+RAG, we set $w_q=0.6$ and $w_v=0.4$.

For experiments on the GQA dataset and the GQA-MSCG dataset,
we fine-tune CFR, Qwen-VL, CFR+RAG, and Qwen-VL+RAG using the train\_balanced split of the GQA dataset and selected the best-performing model weights on the val\_balanced split of GQA.
Using these model weights, we present the experimental results on the test-dev split of the GQA dataset and all seven test splits of our GQA-MSCG dataset.
The maximum number of epochs for fine-tuning CFR and CFR+RAG was set to 12.
The sampled number $T_q$ and $T_v$ for constructing $\mathcal{D}_q$ and $\mathcal{D}_q$ are set to $8$ and $32$, respectively.
Moreover, we use the scene graph provided by GQA to construct $\mathcal{D}_v$.
The number of aggregated primitive $K_q$ and $K_v$ are set to $4$ and $16$, respectively.
Distinctively, for experiments on the VQA v2 dataset,
we set $T_q=1$, $T_v=32$, $K_q=4$ and $K_v=4$.
For constructing $\mathcal{D}_v$, we use the object categories detected by Faster R-CNN \cite{ren2016faster}, ignoring the object attributes.

\subsection{Multi-Sourced Compositional Generalization Performance}

We evaluate the MSCG ability of VQA models on the proposed GQA-MSCG dataset.
We compare with multimodal large models, including LLaVA-1.5 \cite{liu2023improvedllava} and LLaVA-1.6 \cite{liu2024llavanext}, with the experimental results shown in Table \ref{tab:gqamscg_mscg}.
We can observe that:
(1) As the co-occurrence count of novel composition categories in the test samples increases (Level-1 $\rightarrow$ Level-3), the model's performance gradually decreases.
For example, LLaVA-1.5 has an accuracy of $70.22\%$ in the $LL$ split, $67.04\%$ in the $LL+VV$ split, and $64.14\%$ in the $LL+VV+LV$ split.
(2) Different VQA models exhibit significant differences in their ability to generalize to novel compositions of different modality primitives.
For example, LLaVA-1.5 performs better on the $VV$ split than on the $LL$ split, while Qwen-VL shows the opposite.
(3) Both small VQA models (\textit{e.g.}, CFR) and large VQA models (\textit{e.g.}, Qwen-VL) benefit significantly from the proposed framework, which enhances their generalization ability to multi-sourced novel compositions.

As a result, based on the experimental results, the following conclusions can be drawn:
(1) Existing VQA models still have shortcomings in handling complex compositional questions.
(2) It is necessary to specifically consider the multi-sourced compositional generalization ability for VQA models.
(3) The proposed framework is highly versatile and can be applied to different baseline models, improving their generalization ability to novel compositions of primitives from different modalities.

\begin{table}[tb]
	\small
	\centering
 \caption{Accuracy (\%) on the test-dev split of the GQA dataset.}
 \label{tab:gqa_mscg}
\begin{threeparttable}[b]
	\setlength{\tabcolsep}{0.5mm}{
	\begin{tabular}{clc}
	\noalign{\smallskip}\hline\noalign{\smallskip}
Type & Model & Accuracy \\
\cmidrule(){1-3}
\multirow{1}{*}{Attention-based} & MAC \cite{iclr2018mac} & 52.43 \\
\cmidrule(){1-3}
\multirow{1}{*}{Graph-based} & LCGN \cite{hu2019language} & 55.63 \\
\cmidrule(){1-3}
\multirow{1}{*}{NMN-based} & MMN \cite{chen2021meta} & 59.14 \\
\cmidrule(){1-3}
\multirow{3}{*}{Pretrain-based} & BLIP-2 ($\textup{FlanT5}_{\textup{XXL}}$) \cite{li2023blip} & 44.70 \\
\multirow{3}{*}{(zero-shot)} & MiniGPT-4 \cite{zhu2023minigpt} & 43.50 \\
 & LLaVA-1.5 \cite{liu2023improvedllava} & 61.93 \\
 & LLaVA-1.6 \cite{liu2024llavanext} & 64.26 \\
\cmidrule(){1-3}
\multirow{3.5}{*}{Pretrain-based} & CFR \cite{nguyen2022coarse} & 70.27 \\
\multirow{3.5}{*}{(fine-tuned)} &\textbf{+ RAG (Ours)} & \textbf{71.70} \\
\cmidrule(){2-3}
& Qwen-VL \cite{Qwen-VL} & 54.98 \\
&\textbf{+ RAG (Ours)} & \textbf{56.11} \\
	\noalign{\smallskip}\hline\noalign{\smallskip}
	\end{tabular}}
\end{threeparttable}
\end{table}

\begin{table}[tb]
	\small
	\centering
 \caption{Accuracy (\%) on the val split of the VQA v2 dataset.}
 \label{tab:vqa_mscg}
\begin{threeparttable}[b]
	\setlength{\tabcolsep}{0.5mm}{
	\begin{tabular}{clc}
	\noalign{\smallskip}\hline\noalign{\smallskip}
Type & Model & Accuracy \\
\cmidrule(){1-3}
\multirow{4}{*}{Small Model} & DLR \cite{jing2020overcoming} & 57.96 \\
\multirow{4}{*}{($\leq$ 0.2B)} & CF-VQA \cite{niu2021counterfactual} & 63.73 \\
& CLS \cite{mao2024overcoming} & 63.94 \\
& ASS \cite{li2024adversarial} & 64.00 \\
& KDAR \cite{peng2024overcoming} & 65.54 \\
\cmidrule(){1-3}
\multirow{3.5}{*}{Large Model} & PNP-VQA \cite{tiong2022plug} & 63.30 \\
\multirow{3.5}{*}{($\geq$ 7B)} & BLIP-2 ($\textup{FlanT5}_{\textup{XXL}}$) \cite{li2023blip} & 65.20 \\
\cmidrule(){2-3}
& Qwen-VL \cite{Qwen-VL} & 69.04 \\
&\textbf{+ RAG (Ours)} & \textbf{69.82} \\
	\noalign{\smallskip}\hline\noalign{\smallskip}
	\end{tabular}}
\end{threeparttable}
\end{table}

\subsection{Independent and Identically Distributed Generalization Performance}

We verify the improvement of the proposed framework in IID generalization performance on the test-dev split of the GQA dataset \cite{cvpr2019gqa} and the val split of the VQA v2 dataset \cite{goyal2017making}.
For GQA, we compare to five different types of VQA models, including attention-based MAC \cite{iclr2018mac}, graph-based LCGN \cite{hu2019language}, neural modular network based (NMN-based) MMN \cite{chen2021meta}, zero-shot pretrain-based BLIP-2 \cite{li2023blip}, MiniGPT-4 \cite{zhu2023minigpt}, LLaVA-1.5 \cite{liu2023llava}, LLaVA-1.6 \cite{liu2023improvedllava}, and fine-tuned pretrain-based CFR \cite{nguyen2022coarse}, Qwen-VL \cite{Qwen-VL}.
For VQA v2, we compare to VQA models with less than 0.3B parameters (DLR \cite{jing2020overcoming}, CF-VQA \cite{niu2021counterfactual}, CLS \cite{mao2024overcoming}, ASS \cite{li2024adversarial}, KDAR \cite{peng2024overcoming}) and VQA models with more than 7B parameters (PNP-VQA \cite{tiong2022plug}, BLIP-2 \cite{li2023blip}, Qwen-VL \cite{Qwen-VL}).

Experimental results on the test-dev split of the GQA dataset are shown in Table \ref{tab:gqa_mscg}.
From the table, we can observe that:
(1) Pretrain-based models outperform other types of models (such as attention-based, graph-based, and NMN-based frameworks) in terms of IID generalization performance.
(2) The proposed framework further enhances the IID generalization performance of baseline models (\textit{e.g.}, 1.43\% and 1.13\% absolute performance gains
in accuracy for CFR and Qwen-VL, respectively).
These experimental results demonstrate that the proposed framework is applicable to different task scenarios, including multi-sourced compositional generalization scenario and independent and identically distributed generalization scenarios.

Experimental results on the val split of the VQA v2 dataset are shown in the Table \ref{tab:vqa_mscg}.
The results demonstrate that our framework is inoffensive for IID generalization.
The reason why the performance gains of the proposed framework on VQA v2 are less than on GQA is that the questions in GQA are more compositional and thus are more suitable to be improved by our framework.
Such experimental results further prove the effectiveness of the proposed framework for improving the IID generalization ability of VQA models.

\subsection{Ablation Studies}

The results of ablation studies on the GQA-MSCG dataset using CFR as the baseline model are shown in Table \ref{tab:ablation_mscg}.
From the experimental results, the following conclusions can be drawn:
(1) Conducting retrieval only on the question retrieval database $\mathcal{D}_q$ or the image retrieval database $\mathcal{D}_v$ during training can also improve the baseline model's performance, especially for novel compositions of specific modal primitives. For example, when retrieval is performed only on $\mathcal{D}_v$, improvements are more evident for compositions of visual modal primitives (the \textit{VV} split).
(2) Overall, using both the question and image retrieval databases simultaneously results in the best performance (highest average accuracy across all splits).
These ablation study results confirm the necessity of multi-sourced retrieval during training and demonstrate that the linguistic primitive database $\mathcal{D}_q$ or visual primitive database $\mathcal{D}_v$ are complementary, helping to enhance the baseline model’s generalization ability across different compositions.

\begin{table}[tb]
	\small
	\centering
\caption{Ablation studies on the GQA-MSCG dataset.}
\label{tab:ablation_mscg}
	\setlength{\tabcolsep}{1.5mm}{
	\begin{tabular}{lccccc}
	\noalign{\smallskip}\hline\noalign{\smallskip}
\multirow{2.5}{*}{Model} & \multirow{2.5}{*}{$\mathcal{D}_q$} & \multirow{2.5}{*}{$\mathcal{D}_v$} & \multicolumn{3}{c}{Level-1} \\
\cmidrule(lr){4-6}
 &&& \textit{LL} & \textit{VV} & \textit{LV} \\
\cmidrule(){1-6}
CFR \cite{nguyen2022coarse} & - & - & 74.78 & 72.18 & 73.36 \\
\cmidrule(){1-6}
\multirow{2}{*}{\textbf{+ RAG}} & $\checkmark$ & - & 76.18 & 73.46 & 74.24 \\
\multirow{2}{*}{\textbf{(Ours)}} & - & $\checkmark$ & 75.66 & 73.66 & 74.80\\
 & $\checkmark$ & $\checkmark$ & \textbf{76.28} & \textbf{73.88} & \textbf{75.64} \\
	\noalign{\smallskip}\hline\noalign{\smallskip}
	\end{tabular}}
\end{table}

\begin{figure*}[!h]
	\centering
	\subfigure[Experimental results with \( w_q = w_v \).]{
	\includegraphics[width=0.32\linewidth]{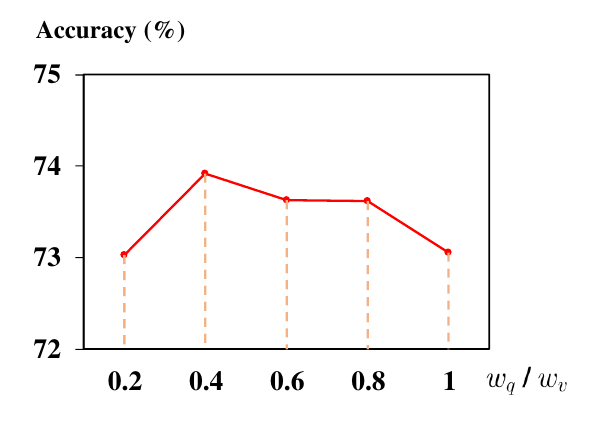}
	}
	\subfigure[Experimental results with \( w_q = 0.4 \) and varying \( w_v \).]{
	\includegraphics[width=0.32\linewidth]{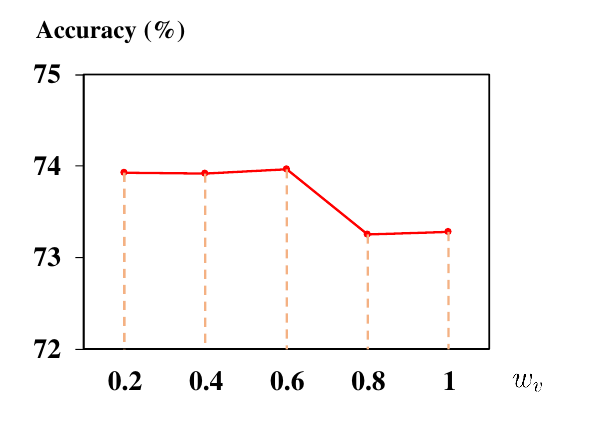}
	}
	\subfigure[Experimental results with \( w_v = 0.4 \) and varying \( w_q \).]{
	\includegraphics[width=0.32\linewidth]{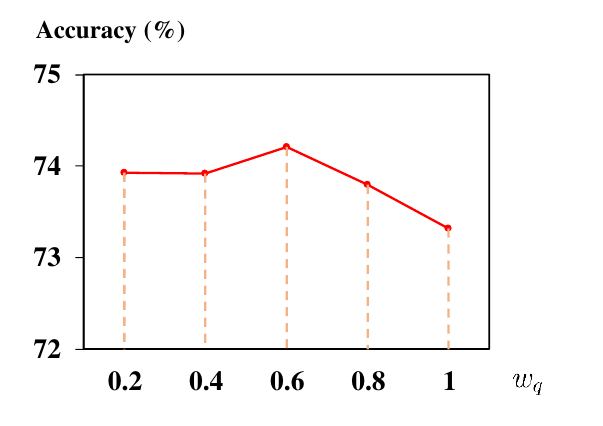}
	}
	\centering
\caption{Parameter analysis using CFR as the baseline model on the GQA-MSCG dataset.}
\label{fig:parameter_mscg}
\end{figure*}

\begin{figure}[tb] 
	\centering
	\includegraphics[width=0.96\linewidth]{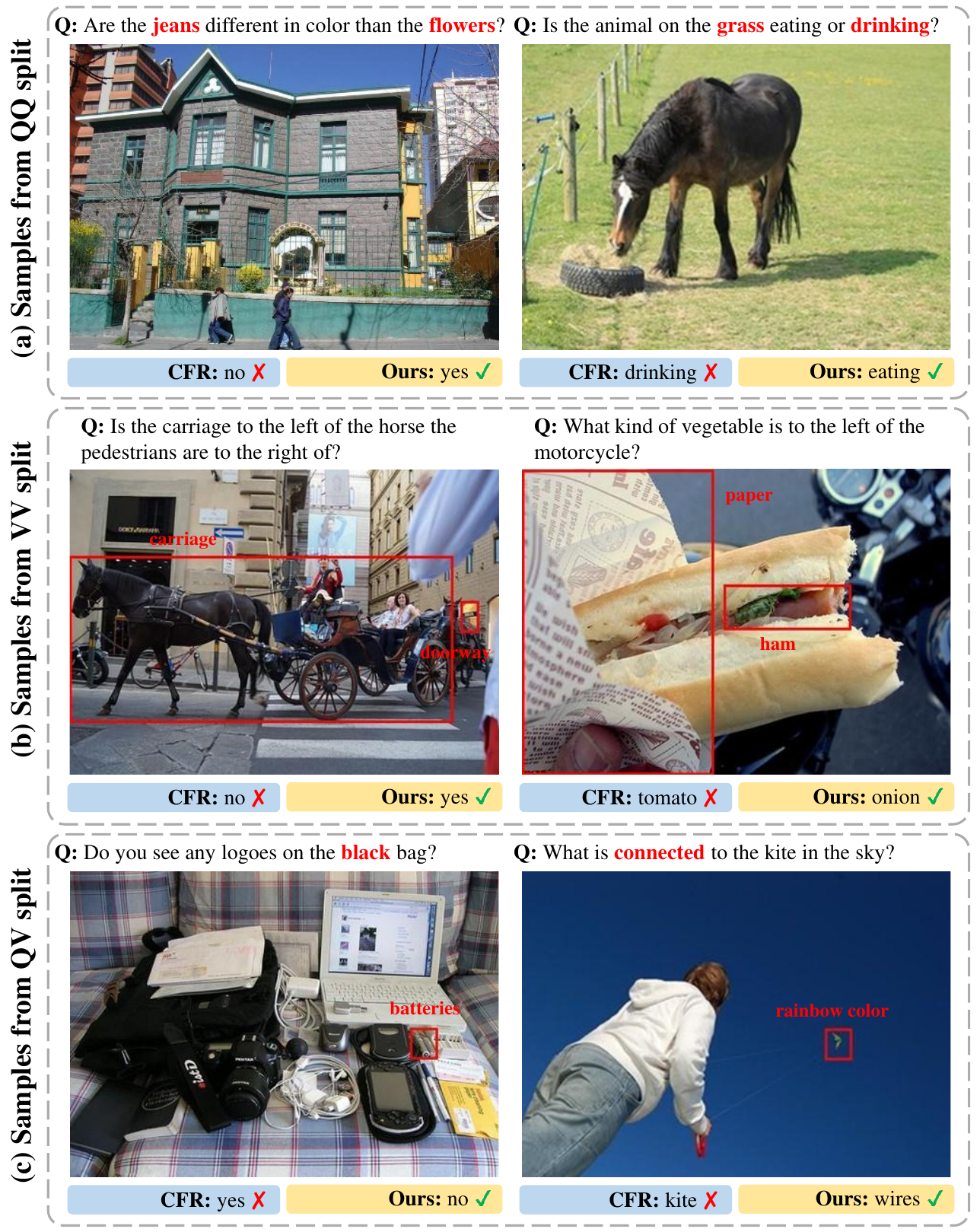}
	\caption{The qualitative comparison between CFR+RAG (Ours) and CFR on the GQA-MSCG dataset.}
	\label{fig:qualitative_mscg}
\end{figure}

\subsection{Parameter Analysis}

We analyze the influences of $w_q$ and $w_v$ on the MSCG ability of our framework, which denote the explicit contributions of the linguistic primitive database and the visual primitive database, respectively. 
Figure \ref{fig:parameter_mscg} shows the performance variations of the proposed framework with changing values of $w_q$ and $w_v$.
First, we conduct an initial analysis of $w_q$ and $w_v$ by setting them to the same value.
The experimental results are shown in Figure \ref{fig:parameter_mscg} (a).
It can be seen that as the values increase, the performance of the proposed framework peaks when $w_q = w_v = 0.4$, and further increasing $w_q$ and $w_v$ does not improve the performance.
Furthermore, we analyze the effectiveness of the proposed framework with $w_q \neq w_v$ by fixing $w_q$ or $w_v$ at $0.4$ and adjusting the other parameter (\textit{i.e.}, $w_v$ or $w_q$), as shown in Figures \ref{fig:parameter_mscg} (b) and (c).
We can observe that:
(1) The performance of CFG+RAG fluctuates obviously when adjusting either $w_q$ or $w_v$.
(2) CFR+RAG performs best with setting $w_q=0.6$ and $w_v=0.4$ simultaneously.
Based on the above experimental results, we set $w_q$ to $0.6$ and $w_v$ to $0.4$ for all experiments.

\subsection{Qualitative Analysis}

For each split of Level-$1$, we provide two qualitative examples of CFR+RAG (Ours) and CFR (the baseline model) on the GQA-MSCG dataset in Figure \ref{fig:qualitative_mscg}.
In the figure, novel compositions of linguistic primitives are highlighted in red text, and visual primitives are enclosed in red boxes.
It can be observed that for test samples with novel compositions composed of primitives from different modalities, the proposed framework helps the baseline model make more accurate predictions.
For example, in the first sample from the GQA-MSCG dataset's $LL$ split, the question is ``Are the jeans different in color than the flowers?", which includes the novel composition of linguistic primitives ``jeans + flowers".
The baseline model CFR gives an incorrect answer, ``no".
In contrast, after incorporating CFR into the proposed framework, it correctly predicts ``yes".
Moreover, for test samples in the $VV$ and $LV$ splits, the proposed framework still provides accurate answers, thanks to the alignment of primitives with the same semantics across different contexts and modalities.
These qualitative examples demonstrate that the proposed framework enhances the baseline model's generalization ability to multi-sourced novel compositions, proving the effectiveness of the framework.

\section{Conclusion}

In this paper, we explored the multi-sourced compositional generalization ability of models in the context of VQA.
We have presented a retrieval-augmented training framework to encourage VQA models to learn unified representations for the same semantic compositions  by aligning semantically equivalent primitives across different modalities at the feature level.
The proposed framework can be seamlessly incorporated into existing VQA models to improve their multi-sourced compositional generalization ability.
We extend the GQA dataset to construct a GQA-MSCG dataset, which enables the quantitative evaluation of the multi-sourced compositional generalization ability for VQA models.
Experimental results demonstrate that our framework  can improve not only the multi-sourced compositional generalization ability, but also the IID generalization ability.

\section*{Acknowledgements}

This work was supported by the Shenzhen Science and Technology Program under Grant No. JCYJ20241202130548062, the Natural Science Foundation of Shenzhen under Grant No. JCYJ20230807142703006, and the Natural Science Foundation of China (NSFC) under Grants No. 62176021 and No. 6217204.

\bibliographystyle{named}
\bibliography{ijcai25}

\end{document}